# Wisdom Of The (AI) Crowd: Investigating Artificial Swarm Intelligence In Large Language Models

Justin Brenne
*Ruhr-University Bochum*, justin.brenne@ruhr-uni-bochum.de

Christian Meske
*Ruhr-Universitaet Bochum*, christian.meske@rub.de

# WISDOM OF THE (AI) CROWD: INVESTIGATING ARTIFICIAL SWARM INTELLIGENCE IN LARGE LANGUAGE MODELS

*Completed Research Paper*

Justin Brenne, Ruhr-University Bochum, Bochum, Germany, Justin.Brenne@rub.de

Christian Meske, Ruhr-University Bochum, Bochum, Germany, Christian.Meske@rub.de

## Abstract

*Human swarm intelligence demonstrates remarkable collective accuracy but faces scalability constraints in cost, coordination, and time. We investigate whether large language models (LLMs) can approximate swarm intelligence effects through artificial swarms, addressing a critical gap in understanding AI-based aggregation mechanisms. We conducted a controlled experiment with 960 manually executed prompts across three proprietary models (GPT-5, Gemini 2.5 Pro, Claude Sonnet 4.5), testing intra-model sampling and inter-model aggregation on eight estimation tasks. Results reveal consistent error reduction through intra- and inter-model aggregation, with significant error reductions up to 37 percentage points in MAPE across different aggregation strategies. We observed small to large effect sizes for positive correlations (Spearman's $\rho=0.242$-$0.568$, all $p<0.001$) between relative confidence interval widths and relative estimation errors, suggesting LLMs possess metacognitive awareness when assessing uncertainty. We discuss implications for research and practice, providing actionable insights for deploying LLM swarms in organizational decision-making.*

## 1 Introduction

The concept of swarm intelligence highlights the ability of groups of individuals to collectively generate more accurate judgments than most of their members could achieve in isolation (Galesic et al., 2023, Willcox et al., 2020). This "wisdom of the crowd" (WoC) has been empirically demonstrated as early as 1906 through Francis Galton's famous ox-weighing contest (Galton, 1907). Since then, it has been shown across diverse domains such as market forecasting (Dong et al., 2024; Schumann et al., 2019), sports betting (Rosenberg & Willcox, 2018), problem solving (Bang & Frith, 2017; Gupta et al., 2024), fake news detection (Wei et al., 2022b). Yet, despite its proven potential and ongoing research (Guo et al., 2025), swarm intelligence in human groups is constrained by inherent drawbacks (especially in technical applications): it requires assembling sufficiently large and diverse groups, is susceptible to biases, social dynamics, and scales poorly in terms of cost, coordination, communication and time (Bullock et al., 2024; Makanda et al., 2023). These limitations fundamentally constrain the applicability of swarm intelligence across many high-stakes decision-making contexts where timely, scalable, and unbiased insights are essential.

While discriminative artificial intelligence (AI) systems have already been thoroughly investigated under the lens of ensemble learning e.g. in the cases of bagging or boosting in machine or deep learning (Ganaie et al., 2022; Mohammed & Kora, 2023; Taudien et al., 2024), the rise of generative artificial intelligence (genAI) and large language models (LLMs) has opened up new possibilities and brought innovative disruptions to several fields with its abilities to produce human-like natural language and internal attention mechanisms (Bandi et al., 2023; Storey et al., 2025; Vaswani et al., 2017). Like in classical ensemble learning, LLM outputs can also be aggregated and combined in different forms, e.g. by averaging results over multiple runs. GenAI thereby provides not only a technical foundation for mimicking swarm intelligence, but also a novel opportunity to extend and stress-test its core principles in artificial systems operating at a scale and speed unattainable by human collectives. Unlike human crowds, LLM systems can be sampled repeatedly at scale, producing diverse outputs through probabilistic variation or by combining different models trained on distinct data. This scalability offers a low-cost, controllable, and repeatable way of approximating "artificial LLM swarms". At the same time, LLMs provides access to vast knowledge bases embedded in training data and has already proven its capacity to perform estimation, prediction, marketing and reasoning tasks (Bubeck et al., 2023; Kshetri et al., 2024). While genAI introduces its own limitations, such as hallucinations (Huang et al., 2025; Zhang et al., 2025), bias or lack of transparency (Bandi et al., 2023), the analogy to swarm intelligence provides a promising avenue to systematically test whether the aggregation of LLM outputs can improve decision quality in ways comparable to, or even beyond, human crowds. The recent proliferation of diverse, accessible genAI models with a variety of features (e.g. Bubeck et al., 2023) now makes systematic investigation of artificial swarms both feasible and timely. The need is especially urgent as an increasing number of organizations deploy genAI for decision-support tasks (Albashrawi, 2025). Understanding whether and how aggregation strategies can improve output reliability has therefore become both theoretically interesting and practically urgent, with recent perspectives beginning to highlight how LLMs can reshape collective intelligence (Burton et al., 2024).

Despite extensive research on human swarm intelligence and discriminative AI ensembles, a profound gap exists at their intersection: peer-reviewed information systems (IS) literature contains limited systematic investigation of whether LLMs can approximate swarm intelligence effects. We address this gap by exploring LLM Swarms, ensembles generated by repeatedly sampling single models or combining multiple models of varying architecture and scale. We examine whether such swarms reduce estimation errors on tasks difficult enough to resist trivial lookup while maintaining verifiable ground truth. We investigate aggregation methods including intra-model sampling and inter-model variance, focusing on numerical estimations and self-reported confidence intervals. We investigate the following central research questions:

- **RQ1**: To what extent does aggregating multiple estimations from the same large language model reduce error on uncertain tasks compared to individual model outputs?
- **RQ2**: How does combining estimations from multiple large language models influence the accuracy of aggregated estimates compared to single-model aggregation?

To address these questions, we designed a controlled study with three LLMs, representing major proprietary models: GPT 5, Gemini 2.5 Pro and Claude Sonnet 4.5. We compare two modes of aggregation: (1) repeated prompting of single models to capture intra-model variance, and (2) aggregation of all models to assess the overall potential of fully heterogeneous LLM swarms to approximate swarm intelligence. The task set consists of eight estimation challenges carefully balanced between uncertainty and ground truth availability: five Fermi problems (rough, quick estimates of big and uncertain quantities) from a publicly available dataset with given reference solutions, and three anonymized economy-based guessing problems derived from historic financial data.

Our contribution addresses the identified gap and is threefold. First, we introduce LLM Swarms as a novel construct that draws on swarm intelligence principles from human collectives to artificial systems, thereby opening a new research stream at the intersection of collective intelligence and genAI in IS. Second, we provide a proof-of-concept methodology for evaluating LLM swarms on uncertain estimation tasks, establishing a rigorous approach that avoids trivial lookups while maintaining ground truth verification. Third, we deliver first systematic empirical evidence by testing three diverse LLM models across intra-model and inter-model aggregation strategies. This establishes the first systematic investigation of LLM-based swarm intelligence, addressing a critical research gap with immediate implications for organizations deploying genAI for decision support. Understanding aggregation strategies' impact on output reliability informs deployment and scaling decisions while theoretically extending swarm intelligence principles to artificial systems, potentially enabling the WoC effect without the traditional constraints.

The remainder of this paper is structured as follows. Section 2 reviews theoretical background and related work on swarm intelligence and emerging genAI Swarms. Section 3 details our methodology, including experimental design, data collection, and analytical measures. Section 4 presents empirical results, followed by Section 5, which situates findings within broader literature and discusses theoretical and practical implications. Section 6 concludes our work by summarizing contributions and addressing limitations.

# 2 Theoretical Background and Related Work

## 2.1 Swarm Intelligence

Based on Willcox et al. (2020, p.1), swarm intelligence is a "natural phenomenon in which social organisms amplify their decision-making abilities by forming real-time systems that converge on optimized solutions.". The concept originated in biology, where researchers studied how e.g. different animals locate food, or how they decide on new nest locations, or how birds coordinate flight paths. These "natural systems" demonstrate how, through mechanisms of self-organization and local communication, groups can repeatedly outperform individual agents in solving problems that require exploration, adaptation, and resilience (Galesic et al., 2023; Willcox et al., 2020).

The translation of this principle into the social sciences has been captured by the idea of the wisdom of the crowd. Here, the central claim is that the average of many independent judgments can be more accurate than most individual judgments, and often even outperform expert predictions (Galesic et al., 2023; Rosenberg, 2016). The famous example of Francis Galton's ox-weighing contest in 1906 illustrates this: while individual guesses varied widely, the mean of the crowd's estimations was remarkably close to the true weight (Galton, 1907). Subsequent studies have confirmed this effect across domains such as financial forecasting, political elections, sports betting or fake news detection, where aggregation helps cancel out random errors and mitigate extreme deviations (Dong et al., 2024; Gupta et al., 2024; Rosenberg & Willcox, 2018; Schumann et al., 2019, Wei et al., 2022b). The advantages of group decision-making stem from the ability to overcome individual biases and leverage diversity in searching the hypothesis space (Bang & Frith, 2017; Hong & Page, 2004).

Additionally, investigations on how to improve the swarm intelligence phenomenon are still ongoing as demonstrated by Willcox et al. (2021, 2023) and their idea of "hyperswarm ranking" to further improve the WoC. This logic can be extended recursively: if individual WoC estimates are themselves treated as inputs to a higher-order aggregation process, swarms of swarms can be formed, potentially enabling a progressive densification of collective knowledge that further reduces variance and pushes aggregate estimates closer to ground truth. Past research has already demonstrated that such approaches have the potential to increase accuracy by randomly selecting a swarm subset and aggregating the results (Willcox et al., 2019). A different concept is the idea of "Slider-Swarms", aiming to amplify collective accuracy by aggregating real-time inputs from multiple participants (Domnauer et al., 2022).

Beyond aggregation strategies, recent modeling efforts explore the underlying mechanisms of interaction that produce swarm intelligence. For example, Guo et al. (2025) investigated multi-agent systems modeled group imitation behavior by integrating information dissemination (via a "perception radius") with social mechanics. The findings show that the consistency of group behavior is positively correlated with the intensity of information interaction among individuals, providing a framework for understanding how swarm intelligence emerges from local interactions (Guo et al., 2025). This focus on information diffusion and imitation is critical, as these same social dynamics can also introduce risks. This focus on interaction mechanisms is part of a broader shift in research beyond static collective intelligence towards dynamic "collective adaptation" (Galesic et al., 2023). This perspective analyzes how collectives dynamically alter their strategies and social structures in response to problems, rather than just measuring if they find an optimal solution. Such a framework is vital for understanding why collectives sometimes fail, exhibit "collective myopia", or become locked into suboptimal path dependencies, highlighting the risks of flawed group processes (Galesic et al., 2023).

Despite its promise, human-based swarm intelligence faces persistent practical and structural challenges that limit its broad applicability. Assembling sufficiently diverse groups often requires significant effort, coordination, and cost, particularly when specialized expertise is needed across multiple domains simultaneously, a requirement that becomes increasingly difficult to satisfy as decision problems grow in complexity and time sensitivity. Maintaining independence of responses is difficult in practice, as individuals are influenced by peer opinions, media, or cultural biases (Dissanayake et al., 2015). Moreover, scaling swarm intelligence (on a technical level) to large groups is highly bias-dependent and resource-intensive: organizing, compensating, and coordinating hundreds or thousands of (diverse) participants is costly and time-consuming (Bullock et al., 2024; Makanda et al., 2023). These challenges limit the scalability and applicability of swarm-based methods in fast-moving or resource-constrained decision contexts. Consequently, scholars and practitioners alike are interested in whether technological systems might provide an alternative or complementary means of harnessing swarm intelligence.

## 2.2 Generative AI Swarms and Multi-Agent LLM Systems

Despite the conceptual appeal of the swarm intelligence and genAI intersection, a substantial gap exists in IS literature regarding LLM swarms. While IS publications address multi-agent systems incorporating multiple LLMs, existing work predominantly focuses on task delegation and agent coordination rather than the aggregation of independent estimations to improve collective accuracy, the core mechanism underlying swarm intelligence. Our literature review of IS research found limited empirical evidence on whether LLMs can replicate the WoC phenomenon observed in human collectives. This section therefore reviews related research on LLM orchestrations and multi-agent LLM systems.

GenAI represents a class of machine learning technologies capable of producing novel content such as text, images, audio, or code (e.g. Storey et al., 2025). Its current prominence stems largely from large language models (LLMs) built on transformer architectures and relies on attention mechanism capabilities (Vaswani et al., 2017). Models such as GPT, Claude, Gemini, and Llama demonstrate impressive proficiency in summarizing information, drafting arguments, solving problems, and even engaging in domain-specific reasoning tasks (Bandi et al., 2023; Bubeck et al., 2023; Feuerriegel et al., 2023; Kshetri et al., 2024).

A critical enabler for LLM swarms is the stochastic nature of language models, where outputs are sampled from probability distributions rather than being deterministic (Bandi et al., 2023). This feature enables simulation of multiple quasi-independent judgments from a single model, mimicking the diversity of opinions in human groups. Additionally, the availability of multiple models trained by different organizations on different data sources adds another layer of diversity (Storey et al., 2025), potentially allowing artificial swarms to capture varied perspectives analogous to human swarm intelligence. However, LLM swarms face distinct limitations compared to human collectives. Models are prone to hallucinations, generating confident yet incorrect statements (Huang et al., 2025; Zhang et al., 2025), and biases embedded in training data can resurface in outputs.

The idea of genAI Swarms combines principles from swarm intelligence with the emerging capabilities of genAI. In this context, repeated outputs from the same model, or aggregated outputs from different models (e.g. Lee et al., 2024), are treated as analogous to individual judgments in human groups. By aggregating across these artificial agents, one can investigate whether similar improvements in accuracy, robustness, or creativity emerge. Related work has explored the concept of repeated prompting of LLMs to generate multiple outputs for the same input, rather than relying on a single deterministic answer. Lee et al. (2024) found that exposing users to multiple, potentially inconsistent LLM-generated responses can improve comprehension, despite the presence of conflicting or erroneous outputs. These findings suggest that leveraging the stochastic variability of LLMs, by prompting the model multiple times, can be a valuable design strategy, highlighting the benefits of "swarm-like" generation approaches in human-AI decision-support contexts. This concept opens a new line of inquiry at the intersection of AI and collective intelligence, where human limitations may be complemented, or challenged, by genAI swarms. Similarly, Henkenjohann and Trenz (2024) show that interaction patterns between humans and AI affect both outcome quality and perceived responsibility, highlighting the importance of workflow design in human-AI configurations.

Several features make genAI swarms particularly compelling as a scalable and controllable alternative to human-based swarm intelligence, addressing many of the structural limitations outlined above. First, they overcome major constraints of human swarms by providing scalability and resource-efficiency (Bullock et al., 2024; Makanda et al., 2023): models can be queried as often as needed, generating hundreds of samples without logistical hurdles. Second, they allow researchers to experiment systematically with aggregation mechanisms or prompts, testing whether averages, medians, or trimmed means yield better approximations of ground truth across different task domains (e.g. Lyon et al., 2015). Third, the heterogeneity of available models provides opportunities for cross-model aggregation. By combining outputs from models of varying sizes, architectures, and training data, researchers can replicate the diversity of perspectives that underpins the wisdom of the crowd.

Recent advances in LLM-based multi-agent systems have demonstrated that coordinating multiple AI agents can enhance problem-solving capabilities beyond single-model approaches. Li et al. (2024) provide a comprehensive framework for LLM-based multi-agent systems, identifying five key components: profile, perception, self-action, mutual interaction, and evolution, that enable collaborative problem-solving. Göldi and Rietsche (2024) distinguish between simple chained model calls and sophisticated agent systems with iterative reasoning and external tools.

Practical applications demonstrate promising results across domains. Khan et al. (2025) achieved 94% accuracy in a two-reviewer workflow using GPT-4-turbo and Claude-3-Opus collaboration, while Gonnermann-Müller et al. (2025) developed FACET for personalized educational materials through specialized agent roles. Analogous to swarm intelligence, these LLM-based agents interact through specialized knowledge domains and distinct functional roles to aggregate information collectively.

Fischer-Brandies et al. (2024) further illustrated how LLM-based agent systems can augment both divergent and convergent thinking in ideation processes, with expert evaluations confirming the artifact's usefulness for integration into existing workflows. The design of human-AI collaboration interfaces also influences performance outcomes. Memmert & Tavanapour (2023) and Memmert (2024) found that LLM-assisted brainstorming can stimulate cognitive processes similar to human groups, though interface design critically affects performance outcomes.

From a swarm intelligence perspective, Burton et al. (2024) specifically address how LLMs can reshape traditional swarm intelligence mechanisms, noting both opportunities for enhanced aggregation and coordination alongside risks related to information access and transmission. While their focus remains on human-AI hybrid systems, they emphasize the need to examine how LLMs affect humans' collective problem-solving capabilities. Furthermore, Burton et al. (2024, p.1645) argue that LLM's "(...) can also be viewed as a product of CI [Collective Intelligence]. LLMs are trained on collective data that encapsulate the contributions of countless individuals (...). Prompting an LLM with a question is like a distilled form of crowdsourcing.". Complementing this, Mienye and Swart (2025) provide a comprehensive survey of ensemble LLM methods, demonstrating how combining multiple models can address limitations such as biases and limited interpretability while improving robustness and generalizability, principles that align closely with traditional swarm intelligence objectives.

# 3 Methodology

## 3.1 Experimental Design

To investigate whether swarm intelligence effects can be replicated through LLM-based systems, we designed a controlled experiment with estimation tasks following a within-subject design where multiple LLM's are tested on the same tasks, allowing comparisons both within single models (intra-model variance) and across different models (inter-model diversity). The task set consists of eight estimation problems chosen to balance uncertainty and ground truth: tasks with readily available solutions would trigger factual recall rather than probabilistic estimation, while overly open-ended tasks would lack verifiable ground truth. five tasks are fermi problems from a publicly available dataset requiring rough quantitative estimates (Kalyan et al., 2021). The other three are economy-related problems based on historical data (e.g., stock prices, company performance measures) with anonymized ground truth values to prevent lookup effects. This balanced task mix ensures both methodological rigor and practical relevance.

GPT 5, Gemini 2.5 Pro and Claude Sonnet 4.5 were included as proprietary models and each was queried via their respective web interface (with additional prompting for revisions done via respective web APIs and standard parameters). Data collection used two aggregation modes: A (intra-model, A1-A3) prompted each model 30 times independently, and B (cross-model, B1) combined responses across all six models together (10 runs per model). Each mode totaled 30 prompts per task, enabling comparison of accuracy benefits from repeated sampling versus heterogeneous model combinations, resulting in a total of 960 manually executed and evaluated prompts. Table 1 summarizes utilized aggregation modes and provides a comprehensive overview of used abbreviations for the aggregation modes in the remainder of this paper.

| **Aggregation Mode** | **Involved Models** | **Runs per Task** |
|---|---|---|
| A1 | GPT 5 | 30 |
| A2 | Gemini 2.5 Pro | 30 |
| A3 | Claude Sonnet 4.5 | 30 |
| B1 | GPT 5, Gemini 2.5 Pro, Claude Sonnet 4.5 | 30 runs aggregated (10 per model) |

*Table 1. Overview of utilized aggregation modes, number of runs and involved models.*

## 3.2 Tasks, Prompt Templates and Metrics

The study employed eight estimation tasks designed to balance difficulty, verifiability, and resistance to simple lookup. Five tasks were Fermi problems adapted from Kalyan et al. (2021), requiring order-of-magnitude reasoning about real-world quantities. We initially investigated common swarm intelligence estimation problems such as historical event dates, celebrity ages, or country-specific per-capita GDP (e.g. Simoiu et al., 2019). However, LLMs either retrieved answers directly from training data or performed web lookups, making inferences trivial and homogeneous, contrary to swarm intelligence principles. We therefore selected tasks less likely to be directly retrievable from training data, namely Fermi problems requiring order-of-magnitude reasoning and economic forecasting tasks involving inherent future uncertainty.

These included estimating the average yearly income globally (F1, ground truth: $11428), the cost to rebuild the NYC subway system from scratch (F2, ground truth: $4.5e11), per-capita restaurant spending relative to the federal budget deficit (F3, ground truth: $2796), gasoline usage during an automobile's lifetime (F4, ground truth: 19200L), and blood units donated daily (F5, ground truth: 40700). The remaining three tasks were economic estimation problems based on anonymized historical data. Participants estimated the 2024 average stock price for a digital services company given 2010-2023 data (Eco1, ground truth: $162.88), the 2024 average stock price for an online retail company given similar historical trends (Eco2, ground truth: $184.63), and the Q4-2024 subscriber count for a streaming entertainment company given quarterly data from 2021-2024 (Eco3, ground truth: 158.6e5). Company identities were withheld and given values rounded to prevent models from accessing memorized financial data, ensuring estimates reflected reasoning rather than retrieval. Eco1 and Eco2 stocks were based on Google and Amazon respectively, while Eco3 was based on Disney+ subscribers. Ground truth values were obtained from public financial records for the economic tasks and validated solutions from Kalyan et al. (2021) for Fermi problems. This task design ensures both methodological rigor through verifiable answers and practical relevance by simulating real-world forecasting challenges where perfect information is unavailable. All prompts followed a standardized template to ensure consistency across

models. To enable systematic analysis, models were instructed to provide both a point estimate and a self-reported 90% confidence interval representing its subjective uncertainty range (i.e. the interval within it believes the true value lies with approximately 90% probability). These intervals are not statistical confidence intervals derived from sampling but rather reflect the model's internal uncertainty assessments. Additionally, models were prompted to refrain from outputting any form of reasoning or explanation. This format ensured comparability, facilitated automated parsing, ease of evaluation, and alignment with the principle of swarm aggregation, where only the final judgments matter. When models offered different response modes (e.g. extended reasoning), they were permitted to select these freely.

Two distinct prompt templates were developed to match the nature of each task type while maintaining structural consistency. For Fermi problems, the prompt emphasized the inherent difficulty and uncertainty: "You will be given a difficult estimation question (a Fermi problem). You will not be able to provide a perfectly accurate answer. Your task is to give your best single-number estimate together with a 90% confidence interval. Do not explain, justify, or add text. You may reason internally as you like to reach the answer, just try to get as close as possible by any means. Output only the estimation as a number and the 90% confidence interval. Do not output any reasoning or text, just the numbers." For economic problems, the template was similar: "You will be given an estimation problem. Your task is to give your best single-number estimate together with a 90% confidence interval. Do not explain, justify, or add text. You may reason internally as you like to reach the answer, just try to get as close as possible by any means. Output only the estimation as a number and the 90% confidence interval. Do not output any reasoning or text, just the numbers." Both templates explicitly required structured numerical output consisting of a point estimate and interval bounds, enabling automated parsing while the explicit permission for internal reasoning acknowledged that models might employ chain-of-thought processes internally without exposing them in outputs.

Model outputs were aggregated in Microsoft Excel and subsequently analyzed using Python with the Pandas, NumPy, and SciPy libraries to compute the relevant metrics. For each aggregation mode we calculated the median and the standard deviation with Bessel correction to account for sampling. To evaluate aggregation quality, we computed three complementary metrics. First, mean absolute error (MAE) compares the MAE of individual estimates against the MAE of the aggregated (mean) estimate, where lower values for the aggregated estimate indicate swarm benefits. Second, mean absolute percentage error (MAPE) contrasts the MAPE of individual predictions with the MAPE of the aggregate, offering an interpretable percentage-based accuracy measure. Confidence interval calibration assesses whether confidence intervals are well-calibrated by presenting the proportion of ground truth values falling within individual model intervals. For aggregated intervals, we use the average of lower and upper bounds respectively and check if the ground truth value is contained. Lastly, we examined the correlation between interval width and MAE by calculating Spearman's rank correlation coefficient. We used Spearman's method rather than Pearson's because it is robust to non-linear relationships, extreme value differences across tasks and outliers, making rank-based correlation more suitable for our heterogeneous estimation problems.

# 4 Results

The following Tables 2 to 5 report aggregation performance across all aggregation modes. Table 6 reports the Spearman correlation analysis between the relative width of the confidence intervals and the relative estimation error. Relative interval width refers to the size of the confidence interval expressed as a proportion of the ground truth value, and relative error indicates how far the point estimate deviates from the ground truth, also expressed as a proportion of the ground truth. Effect sizes for correlations are interpreted according to Cohen (1988). Descriptive metrics include the median and standard deviation (SD) of point estimates across runs. Error metrics compare average individual model error (Ind.) against aggregated estimate error (Agg.) using MAE and MAPE. Confidence interval calibration is assessed through two measures: Calib. (%) reports the percentage of individual 90% confidence intervals containing GT, while "GT in Agg. Interval" indicates whether the GT falls within the aggregated interval formed by averaging all lower and upper bounds (reported as True/False). Tasks are labeled F1-F5 for Fermi estimation problems and Eco1-Eco3 for economic estimation problems as defined before.

Table 2 specifically shows the mentioned metrics for aggregation mode A1 – GPT 5. MAE/MAPE benefits can be found for all eight tasks, with MAPE reductions ranging from about 0.43% (Eco3) to 10.68% (F5). Interval calibration ranged from 36.67% to 100% with six out of eight averaged intervals capturing GT.

| Task | Descriptive Metrics | | MAE | | MAPE (%) | | Confidence Interval | |
|---|---|---|---|---|---|---|---|---|
| | Median | SD | Ind. | Agg. | Ind. | Agg. | Calib. (%) | GT in Agg. interval |
| F1 | 14000 | 1779.90 | 2144.10 | 1629.57 | 18.76 | 14.26 | 96.67 | True |
| F2 | 9e11 | 8.22e11 | 6.08e11 | 5.68e11 | 135.19 | 126.30 | 50.00 | False |
| F3 | 3250 | 457.62 | 465.70 | 364.50 | 16.66 | 13.04 | 90.00 | True |
| F4 | 25000 | 10740.19 | 8093.33 | 7400 | 42.15 | 38.54 | 86.67 | True |
| F5 | 34500 | 8270.15 | 8180 | 3833.33 | 20.10 | 9.42 | 80.00 | True |
| Eco1 | 127.50 | 22.36 | 28.59 | 24.48 | 17.55 | 15.03 | 36.67 | True |
| Eco2 | 150.50 | 24.43 | 32.70 | 29.30 | 17.71 | 15.87 | 50.00 | False |
| Eco3 | 1.58e11 | 1.50e6 | 1.42e6 | 7.40e5 | 0.90 | 0.47 | 100.00 | True |

*Table 2. Results of aggregation mode A1 – GPT 5.*

Table 3 specifically shows the mentioned metrics for aggregation mode A2 – Gemini 2.5 Pro. MAE/MAPE benefits can be found for all eight tasks, with MAPE reductions ranging from about 0.35% (Eco2) to 17.58% (F2). Interval calibration ranged from 23.33% to 86.67% with five out of eight averaged intervals capturing GT.

| Task | Descriptive Metrics | | MAE | | MAPE (%) | | Confidence Interval | |
|---|---|---|---|---|---|---|---|---|
| | Median | SD | Ind. | Agg. | Ind. | Agg. | Calib. (%) | GT in Agg. interval |
| F1 | 14200 | 3866.33 | 3267.60 | 2333.47 | 28.59 | 20.42 | 36.67 | True |
| F2 | 1.05e12 | 4.79e12 | 1.89e12 | 1.81e12 | 420.64 | 403.06 | 30.00 | False |
| F3 | 2825 | 816.57 | 629.53 | 169.93 | 22.52 | 6.08 | 56.67 | True |
| F4 | 20000 | 8142.78 | 4530 | 2516.67 | 23.59 | 13.11 | 86.67 | True |
| F5 | 40000 | 10549.50 | 6986.67 | 433.33 | 17.17 | 1.06 | 80.00 | True |
| Eco1 | 122 | 32.70 | 37.62 | 28.78 | 23.10 | 17.67 | 23.33 | False |
| Eco2 | 130.50 | 20.64 | 47.81 | 47.16 | 25.90 | 25.55 | 30.00 | False |
| Eco3 | 1.57e8 | 5.19e6 | 3.58e6 | 3.67e5 | 2.26 | 0.23 | 86.67 | True |

*Table 3. Results of aggregation mode A2 – Gemini 2.5 Pro.*

Table 4 specifically shows the mentioned metrics for aggregation mode A3 – Claude Sonnet 4.5. MAE/MAPE benefits can be found for seven out of eight tasks, with MAPE reductions ranging from about 0% (Eco2) to 9.93% (F5). Interval calibration ranged from 50% to 96.67% with six out of eight averaged intervals capturing GT.

| Task | Descriptive Metrics | | MAE | | MAPE (%) | | Confidence Interval | |
|---|---|---|---|---|---|---|---|---|
| | **Median** | **SD** | **Ind.** | **Agg.** | **Ind.** | **Agg.** | **Calib. (%)** | **GT in Agg. interval** |
| F1 | 12000 | 1819.47 | 1483.87 | 664.13 | 12.98 | 5.81 | 83.33 | True |
| F2 | 2.78e11 | 9.06e10 | 1.87e11 | 1.83e11 | 41.56 | 40.67 | 53.33 | False |
| F3 | 2100 | 750.83 | 832.80 | 825.33 | 29.79 | 29.52 | 50.00 | False |
| F4 | 37500 | 13837.31 | 16893.33 | 16466.67 | 87.99 | 85.76 | 63.33 | True |
| F5 | 75000 | 20577.86 | 26340 | 22300 | 64.72 | 54.79 | 96.67 | True |
| Eco1 | 135 | 11.94 | 22.66 | 22.51 | 13.91 | 13.82 | 76.67 | True |
| Eco2 | 145 | 6.61 | 41.63 | 41.63 | 22.55 | 22.55 | 56.67 | True |
| Eco3 | 1.56e8 | 6.12e6 | 3.91e6 | 1.30e6 | 2.47 | 0.82 | 50.00 | True |

*Table 4. Results of aggregation mode A3 – Claude Sonnet 4.5.*

Table 5 specifically shows the mentioned metrics for aggregation mode B1 – pooled proprietary models. MAE/MAPE benefits can be found for all eight tasks, with MAPE reductions ranging from about 0.66% (Eco3) to 37.41% (F2). Interval calibration ranged from 43.33% to 86.67% with all eight averaged intervals capturing GT.

| Task | Descriptive Metrics | | MAE | | MAPE (%) | | Confidence Interval | |
|---|---|---|---|---|---|---|---|---|
| | **Median** | **SD** | **Ind.** | **Agg.** | **Ind.** | **Agg.** | **Calib. (%)** | **GT in Agg. interval** |
| F1 | 13100 | 1594.67 | 1866.67 | 1490.67 | 16.33 | 13.04 | 73.33 | True |
| F2 | 5e11 | 6.53e11 | 4.93e11 | 3.25e11 | 109.52 | 72.11 | 43.33 | True |
| F3 | 2800 | 857.19 | 635.27 | 167.67 | 22.72 | 6.00 | 66.67 | True |
| F4 | 22500 | 12874.46 | 9813.33 | 8000 | 51.11 | 41.67 | 76.67 | True |
| F5 | 40000 | 18436.72 | 14093.33 | 6433.33 | 34.63 | 15.81 | 86.67 | True |
| Eco1 | 143.50 | 19.64 | 21.58 | 17.58 | 13.25 | 11.04 | 63.33 | True |
| Eco2 | 147.50 | 23.04 | 31.18 | 27.70 | 16.89 | 15.00 | 73.33 | True |
| Eco3 | 1.58e8 | 3.35e6 | 2.51e6 | 1.47e6 | 1.59 | 0.93 | 83.33 | True |

*Table 5. Results of aggregation mode B1 – proprietary models combined.*

Table 6 presents correlation analysis results across aggregation modes, reporting Spearman's ρ, p-values (all $p < 0.001$), sample sizes (n), and effect size interpretations following Cohen (1988). Sample size ($n = 240$) reflects 30 prompts per task across eight tasks. Due to magnitude differences across tasks, we calculated relative confidence interval widths and relative estimation errors as proportions of GT values to enable meaningful comparison.

| Aggregation Mode | Spearman's ρ | p-value | n | Interpretation |
|---|---|---|---|---|
| A1 – GPT 5 | 0.568 | < 0.001 | 240 | Large Effect Size |
| A2 – Gemini 2.5 Pro | 0.283 | < 0.001 | 240 | Small Effect Size |
| A3 – Claude Sonnet 4.5 | 0.509 | < 0.001 | 240 | Large Effect Size |
| B1 – All Models | 0.466 | < 0.001 | 240 | Medium Effect Size |

*Table 6. Results of correlation analysis for all aggregation modes.*

Across aggregation modes A1-A3, proprietary models demonstrated mostly consistent improvement when aggregating estimates compared to individual predictions, with MAE/MAPE reductions observed in most tasks. Claude Sonnet 4.5 (A3) benefited least from aggregation with marginal error reductions, while GPT-5 (A1) and Gemini 2.5 Pro (A2) exhibited more substantial benefits, with error reductions exceeding 15%. Error reduction magnitude varied across tasks, ranging from negligible improvements of 0% (A3 – Eco2) to larger reductions of about 17% (A2 – F2). Confidence interval calibration ranged widely from 23.33% to 100%, indicating models could generate self-reported intervals sometimes containing ground truth values but consistently fell short of the expected 90% confidence level. GPT-5 (A1) and Claude Sonnet 4.5 (A3) performed best with ground truth falling within their average intervals for six out of eight tasks, compared to Gemini 2.5 Pro (A2) which achieved five out of eight. The heterogeneous aggregation mode B1, combining all three proprietary models, generally achieved competitive or superior performance compared to single-model aggregation, particularly in task F2 where MAPE decreased from 109.52% to 72.11%. MAPE reductions are generally better than singular model aggregations. Interval calibration does not improve with pooled models. However, pooling models (B1) resulted in the GT of every task being included in the respective, averaged interval boundaries, showing an improvement over individually aggregated models. For three tasks (F3, Eco1, Eco2) the MAPE of aggregated models (B1) was minimal overall. F1's and F2's MAPE were best for Claude Sonnet 4.5 (A3), while F4's, F5's and Eco3's MAPE was minimal for Gemini 2.5 Pro (A2). Fermi estimation F2 and economic estimation task Eco3 showed highest values for median and SD due to their contextual nature.

Aggregation modes demonstrated small to large effect sizes for positive correlations between relative confidence interval width and relative error (Spearman's $\rho$ ranging from 0.283 to 0.568, all $p < 0.001$), indicating models assign wider confidence intervals to more uncertain estimates. GPT-5 (A1) exhibited the strongest significant correlation ($\rho = 0.568$), while Gemini 2.5 Pro (A2) showed the weakest ($\rho = 0.283$). The heterogeneous proprietary model aggregation (B1) did not strengthen this correlation but yielded an intermediate value ($\rho = 0.466$).

# 5 Discussion

## 5.1 Swarm Intelligence in Large Language Models

Our empirical investigation into LLM swarms reveals nuanced patterns in how artificial swarm intelligence emerges from model aggregation, with results that both validate and challenge core theoretical assumptions.

The most striking finding concerns the consistent error reduction through aggregation across proprietary models (GPT-5, Gemini 2.5 Pro, Claude Sonnet 4.5), with heterogeneous aggregation mode B1 achieving relative error reductions up to nearly 40% for certain configurations. For instance, task F2 saw MAPE reduction from 109.52% to 72.11%, demonstrating that these models possess sufficient internal variance and complementary biases to benefit from aggregation. This improvement suggests that high-capacity models can generate meaningfully diverse outputs despite drawing from overlapping training distributions.

A critical pattern emerges when examining task-specific performance: Economic forecasting problems (Eco1-3) showed markedly more consistent improvements than Fermi problems (F1-5). Economic tasks, anchored in historical time-series data with clear trends and bounded ranges, provided statistical reference points that enabled models to converge on reasonable estimates. For example, Eco3's subscriber forecasting achieved minimal MAPE of 0.23% for Gemini's aggregated estimate. In contrast, Fermi problems like F2's subway reconstruction cost, which lack concrete anchoring points and require pure order-of-magnitude reasoning, exhibited extreme variance with standard deviations reaching 4.79e12 for Gemini. This suggests that aggregation effectiveness is not universal, but rather contingent on task characteristics where models can internally represent consistent magnitude scales. Problems with statistical anchors, historical patterns, bounded ranges, or trend data, appear fundamentally more tractable for LLM swarms than pure estimation challenges requiring abstract reasoning without reference points.

Importantly, our initial attempts using conventional swarm intelligence benchmarks, such as historical event dates or celebrity ages (e.g. Simoiu et al., 2019), proved unsuitable, as LLMs simply retrieved memorized answers rather than engaging in probabilistic estimation. This highlights a crucial boundary condition: LLM swarms require tasks that balance uncertainty with verifiability, neither trivially solvable through lookup nor completely detached from learnable patterns.

Focusing on RQ1, within-model variance analysis (modes A1-A6) reveals differential consistency patterns with practical implications. Claude Sonnet 4.5 demonstrated remarkably low standard deviations across tasks, for instance, only 6.61 in Eco2 versus Gemini's 20.64, producing near-identical estimates in repeated sampling. While this consistency ensures reliability when the model's reasoning is sound, it amplifies systematic errors when flawed, underscoring the fundamental challenge of lacking true independence in repeated sampling from the same model. This behavior mirrors the "echo chamber" effect observed in human groups where insufficient diversity leads to false confidence.

The heterogeneous aggregation results, being the focus of RQ2, present unexpected complexity that challenges naive "more is better" assumptions. Mode B1's universal success in capturing ground truth within aggregated confidence intervals (6/6 tasks) compared to individual models (4/6 for GPT-5 and Claude, 3/6 for Gemini) demonstrates clear benefits of model diversity. However, this improvement in uncertainty coverage does not always translate to superior point estimates, suggesting that different models contribute complementary perspectives that broaden uncertainty bounds without necessarily improving central tendency.

The aggregation benefits observed across models are likely driven by a combination of mechanisms: independent variance enabling statistical error cancellation, and complementary bias correction, whereby models trained on institutionally and architecturally distinct data sources systematically cover each other's blind spots. This also explains the differential performance between task types: economic forecasting tasks provide statistical anchors in the form of historical trends that models can collectively converge on, whereas Fermi problems require order-of-magnitude reasoning without such reference points, reducing the shared signal that aggregation can amplify.

Consider a concrete example: In task F1 (global average income), individual models showed calibration rates of 96.67% (GPT-5), 36.67% (Gemini), and 83.33% (Claude), yet all achieved similar MAPE reductions through aggregation. This pattern suggests that while models struggle with absolute calibration, they maintain relative accuracy in distinguishing between more and less certain estimates, a form of ordinal metacognition that proves valuable for aggregation despite poor cardinal calibration.

Task-specific analysis reveals that problems requiring creative extrapolation beyond training data (like F2's hypothetical subway reconstruction with its $4.5e11 ground truth) may be inherently more unsuited to LLM swarms due to the absence of learned patterns to guide estimation. Conversely, problems grounded in observable trends (like stock price forecasting in Eco1-2) benefit substantially from aggregation, with MAPE improvements ranging from 2-6 percentage points. This dichotomy suggests a fundamental boundary condition: LLM swarms excel when aggregating over learned statistical regularities but have reduced benefits when confronting novel estimation challenges lacking training data precedents.

The correlation analysis reveals another layer of sophistication in LLM behavior. We observed small-to-large positive correlations ($\rho = 0.283$-$0.568$) between relative confidence interval widths and relative estimation errors across all aggregation modes. This consistent relationship, maintained in both single-model and heterogeneous aggregation, suggests that LLMs possess metacognitive awareness when assessing their own uncertainty. GPT-5 demonstrated the strongest correlation ($\rho = 0.568$), indicating superior self-assessment capabilities, while Gemini Pro 2.5's weaker correlation ($\rho = 0.283$) hints at less developed uncertainty quantification. Intriguingly, the heterogeneous aggregation (B1) yielded an intermediate correlation ($\rho = 0.466$), suggesting that mixing models with varying metacognitive abilities produces average self-awareness rather than enhanced collective metacognition. However, observed correlations between confidence interval width and estimation error may partly reflect task-level difficulty rather than genuine within-task metacognitive awareness. Tasks where models systematically struggle may produce both wider intervals and larger errors simultaneously, inflating the apparent metacognitive signal. While our use of relative metrics as proportions of ground truth mitigates cross-task magnitude differences, it cannot fully disentangle task-difficulty effects from true within-task uncertainty calibration.

## 5.2 Implications for Research and Practice

From a theoretical perspective, our results suggest that LLM swarms can partially replicate human swarm intelligence effects, but through fundamentally different mechanisms. Building on Burton et al.'s (2024) framework of LLMs as products of swarm intelligence, our findings suggest a recursive phenomenon: models trained on collective human knowledge can themselves form collectives that approximate swarm intelligence. This creates what might be termed "second-order swarm intelligence", aggregation of systems that are themselves aggregations of human knowledge.

The strong correlation between confidence interval width and estimation error extends the collective intelligence literature in novel directions. Unlike human swarms where metacognition might vary dramatically across individuals, LLMs demonstrate consistent metacognitive patterns within models. This suggests opportunities to develop new theoretical frameworks that account for artificial metacognition in collective systems. Future research should investigate whether this metacognitive consistency can be leveraged to weight individual contributions dynamically, like Willcox et al.'s (2021) hyperswarm ranking but adapted for artificial agents. Extending current investigations on swarm intelligence strategies (e.g. Galesic et al., 2023), future theoretical perspectives should also consider the idea of multiple LLMs interacting to strategize aggregation actively over time, bridging the gap towards "AI swarm agents".

The task-dependency of aggregation benefits aligns with Hong and Page's (2004) diversity theorem but requires theoretical refinement. While human diversity stems from varied experiences and cognitive styles, LLM diversity emerges from capacity, architectural differences, training data variations, and stochastic sampling. Research should develop new taxonomies of artificial diversity that map these technical variations to collective performance outcomes. Specifically, investigations could explore whether architectural diversity (transformer variants), scale diversity (parameter counts), or training diversity (datasets and objectives) contributes most to swarm effectiveness.

For practitioners, our results offer immediately actionable insights that can guide LLM deployment strategies. Organizations should adopt a task-contingent approach to LLM swarm deployment, matching aggregation strategies to problem characteristics. For forecasting tasks with historical anchors, market predictions, demand planning, trend extrapolation, our results suggest implementing multi-model aggregation with at least three high-capacity proprietary models. The consistent error reduction observed in economic tasks translates directly to improved decision quality in these domains.

However, for creative estimation or problems lacking statistical precedents, organizations should temper expectations. A hybrid approach leveraging Henkenjohann & Trenz's (2024) insights on human-AI collaboration patterns could combine LLM swarms for initial estimates with human expert validation for problems exceeding uncertainty thresholds. Practitioners could operationalize this by using the correlation between confidence intervals and errors: when models report wide intervals (suggesting high uncertainty), human oversight becomes critical. Organizations can implement tiered response systems where narrow confidence intervals trigger automated execution, moderate intervals prompt human review, and wide intervals escalate to expert committees. This approach, potentially extending existing multi-agent frameworks (e.g. Göldi & Rietsche, 2024), creates adaptive workflows that match decision authority to uncertainty levels.

The scalability advantages of LLM swarms, e.g. instant responses, unlimited sampling or absence of coordination costs, make them particularly valuable for time-sensitive decisions. Building on Albashrawi's (2025) insights for genAI in decision-making, organizations could implement rapid prototyping cycles where LLM swarms generate initial estimates for immediate action while human teams conduct deeper analysis in parallel. This temporal decoupling addresses the speed-accuracy tradeoff inherent in traditional decision-making.

The intersection of swarm intelligence and generative AI opens numerous research avenues. Future research should expand the experimental scope to include more models, diverse task categories beyond numerical estimation (e.g. classification, ranking or qualitative judgment tasks), and refined aggregation mechanisms. Beyond static aggregation, future research should explore dynamic LLM swarm agents capable of iterative interaction and debate prior to aggregation, more closely replicating the feedback-driven mechanisms underlying biological swarm intelligence. Extending Lyon et al.'s (2015) work on confidence interval aggregation methods, researchers should investigate whether alternative aggregation mechanisms, trimmed means, weighted averages based on confidence, or Bayesian model averaging, improve upon simple averaging for LLM swarms. The potential for prompt engineering to induce greater output diversity represents another frontier: instructing models to adopt different analytical frameworks or personas might artificially create the perspective diversity that naturally emerges in human crowds.

Future research should also revisit the generalizability of these findings as open-source models continue to close the capability gap with proprietary systems. Frontier-scale open-source releases rivaling GPT, Claude, or Gemini in capacity may exhibit the internal variance and metacognitive signaling necessary to produce meaningful swarm benefits comparable to those observed here.

Integration with existing organizational processes also requires investigation. Following Fischer-Brandies et al.'s (2024) work on LLM-augmented ideation, research should examine how LLM swarms integrate with established decision-making workflows, corporate governance structures, and regulatory requirements. Questions of accountability, explainability, and audit trails become critical when LLM swarms influence high-stakes decisions.

The evidence for differential performance between task types suggests developing task taxonomies that predict aggregation effectiveness. Researchers could build on existing LLM ensemble method surveys (e.g. Mienye & Swart, 2025) to create tools that assess whether a given problem suits LLM swarm approaches based on characteristics like data availability, problem structure or required creativity. Systematic benchmarking of LLM swarms against published human swarm results on comparable tasks represents a particularly promising avenue for future research, as it would allow direct assessment of whether artificial swarms achieve similar, inferior, or superior collective accuracy to human collectives. Developing standardized benchmarks for evaluating artificial swarm intelligence would facilitate cross-study comparisons and accelerate theoretical development.

Finally, the hybrid human-AI swarm represents perhaps the most promising direction. Combining human creativity and contextual understanding with AI's computational scale could yield collective intelligence exceeding either system alone. Research should investigate optimal mixing ratios, interaction protocols, and aggregation methods for these hybrid systems, potentially revolutionizing how organizations approach complex decision-making.

# 6 Conclusion

This study provides the first systematic investigation of whether LLMs can replicate swarm intelligence effects, establishing LLM swarms as a promising avenue for both researchers and practitioners. Our findings provide initial evidence that artificial swarm intelligence through model aggregation offers a scalable, rapid, and cost-effective way of ensemble LLM aggregation, leveraging high-capacity proprietary models. The ability to instantly generate hundreds of estimates without coordination overhead represents a transformative capability for organizational decision-making, especially in time-sensitive contexts where assembling human experts is impractical. Furthermore, we showed that empirical evidence points to some form of metacognition in LLMs, as their self-reported confidence interval-width correlated with the calculated relative error for all aggregation modes.

However, several limitations constrain the generalizability of our findings. The study's proof-of-concept nature, with limited sample sizes and task diversity, provides initial evidence but requires extensive validation across broader problem domains and model architectures. Shared training biases and the potential for multiple identical hallucinations pose risks that may be less prevalent in human swarms with genuine cognitive diversity. Additionally, the black-box nature of LLMs makes it more difficult to trace reasoning paths compared to human deliberation, potentially limiting accountability in high-stakes decisions. The sensitivity of model outputs to prompt formulation also introduces methodological challenges that warrant systematic investigation. Furthermore, the temperature settings of proprietary models are opaque and vary greatly in exposure and functionality between model providers, with even minor adjustments potentially resulting in substantial variation in output. Furthermore, the task set remains constrained to numerical estimation with verifiable ground truth, limiting generalizability to other decision-making contexts such as medical diagnosis or legal forecasting. In addition, correlations should be interpreted with caution, as observed associations do not imply causation.

As generative AI continues to evolve, understanding how to harness artificial swarm intelligence from artificial agents will become increasingly critical for organizations seeking to augment human decision-making capabilities while navigating the inherent limitations of both biological and artificial intelligence systems.

## 7 Acknowledgements

During the preparation of this work the authors used OpenAI's ChatGPT 5 as well as Anthropic's Claude Sonnet 4.5 / Opus 4.1 in order to improve language clarity, grammar and style. After using this tool / service, the authors reviewed and edited the content as needed and take full responsibility for the content of the article.

## References

Albashrawi, M. (2025). Generative AI for decision-making: A multidisciplinary perspective. *Journal of Innovation & Knowledge, 10*(4), 100751. https://doi.org/10.1016/j.jik.2025.100751

Bandi, A., Adapa, P. V. S. R., & Kuchi, Y. E. V. P. K. (2023). The power of generative AI: A review of requirements, models, input–output formats, evaluation metrics, and challenges. *Future Internet, 15*(8), 260. https://doi.org/10.3390/fi15080260

Bang, D., & Frith, C. D. (2017). Making better decisions in groups. *Royal Society Open Science, 4*(8), 170193. https://doi.org/10.1098/rsos.170193

Bubeck, S., Chandrasekaran, V., Eldan, R., Gehrke, J., Horvitz, E., Kamar, E., Lee, P., Lee, Y. T., Li, Y., Lundberg, S., Nori, H., Palangi, H., Ribeiro, M. T., & Zhang, Y. (2023). Sparks of artificial general intelligence: Early experiments with GPT-4 *(arXiv:2303.12712)*. arXiv. https://doi.org/10.48550/arXiv.2303.12712

Bullock, S., Ajmeri, N., Batty, M., Black, M., Cartlidge, J., Challen, R., Chen, C., Chen, J., Condell, J., Danon, L., Dennett, A., Heppenstall, A., Marshall, P., Morgan, P., O'Kane, A., Smith, L. G. E., Smith, T., & Williams, H. T. P. (2024). Artificial intelligence for collective intelligence: A national-scale research strategy. *The Knowledge Engineering Review, 39*, e10. https://doi.org/10.1017/S0269888924000110

Burton, J. W., Lopez-Lopez, E., Hechtlinger, S., Rahwan, Z., Aeschbach, S., Bakker, M. A., Becker, J. A., Berditchevskaia, A., Berger, J., Brinkmann, L., Flek, L., Herzog, S. M., Huang, S., Kapoor, S., Narayanan, A., Nussberger, A.-M., Yasseri, T., Nickl, P., Almaatouq, A., … Hertwig, R. (2024). How large language models can reshape collective intelligence. *Nature Human Behaviour, 8*, 1643–1655. https://doi.org/10.1038/s41562-024-01959-9

Cohen, J. (1988). Statistical power analysis for the behavioral sciences (2nd ed.). *Routledge*. https://doi.org/10.4324/9780203771587

Dissanayake, I., Zhang, J., & Gu, B. (2015). Task division for team success in crowdsourcing contests: Resource allocation and alignment effects. *Journal of Management Information Systems, 32*(2), 8–39. https://doi.org/10.1080/07421222.2015.1068604

Domnauer, C., Willcox, G., & Rosenberg, L. (2022). Collaborative forecasting using "slider-swarms" improves probabilistic accuracy. In *Proceedings of the Future Technologies Conference (FTC) 2022, Volume 1* (pp. 378–392). Lecture Notes in Networks and Systems, 559. Springer. https://doi.org/10.1007/978-3-031-18461-1_26

Feuerriegel, S., Hartmann, J., Janiesch, C., & Zschech, P. (2023). Generative AI. *Business & Information Systems Engineering, 66*(1), 111-126. https://doi.org/10.1007/s12599-023-00834-7

Fischer-Brandies, L., Meierhöfer, S., & Protschky, D. (2024). Augmenting divergent and convergent thinking in the ideation process: An LLM-based agent system. In *Proceedings of the European Conference on Information Systems (ECIS 2024)*. Association for Information Systems. https://aisel.aisnet.org/ecis2024/track20_adoption/track20_adoption/13

Galesic, M., Barkoczi, D., Berdahl, A. M., Biro, D., Carbone, G., Giannoccaro, I., Goldstone, R. L., Gonzalez, C., Kandler, A., Kao, A. B., Kendal, R., Kline, M., Lee, E., Massari, G. F., Mesoudi, A., Olsson, H., Pescetelli, N., Sloman, S. J., Smaldino, P. E., & Stein, D. L. (2023). Beyond collective intelligence: Collective adaptation. *Journal of the Royal Society Interface, 20*(200), 20220736. https://doi.org/10.1098/rsif.2022.0736

Galton, F. (1907). Vox populi. *Nature, 75*, 450–451. https://doi.org/10.1038/075450a0

Ganaie, M. A., Hu, M., Malik, A. K., Tanveer, M., & Suganthan, P. N. (2022). Ensemble deep learning: A review. *Engineering Applications of Artificial Intelligence, 115*, 105151. https://doi.org/10.1016/j.engappai.2022.105151

Gonnermann-Müller, J., Haase, J., Fackeldey, K., & Pokutta, S. (2025). FACET: Teacher-centred LLM-based multi-agent systems—Towards personalized educational worksheets *(arXiv preprint arXiv:2508.11401v2)*. https://arxiv.org/abs/2508.11401

Göldi, A., & Rietsche, R. (2024). Making sense of large language model-based AI agents. In *Proceedings of the International Conference on Information Systems (ICIS 2024)*. Association for Information Systems. https://aisel.aisnet.org/icis2024/aiinbus/aiinbus/16

Guo, C., Yu, P., Li, M., & Chen, X.-B. (2025). Multi-agent imitation behavior based on information interaction. *Complexity, 2025*, Article 8828678. https://doi.org/10.1155/cplx/8828678

Gupta, P., Kim, Y. J., Glikson, E., & Woolley, A. W. (2024). Using digital nudges to enhance collective intelligence in online collaboration: Insights from unexpected outcomes. *MIS Quarterly, 48*(1), 393–408. https://doi.org/10.25300/MISQ/2023/16752

Henkenjohann, R., & Trenz, M. (2024). Challenges in collaboration with generative AI: Interaction patterns, outcome quality and perceived responsibility. In *Proceedings of the 32nd European Conference on Information Systems (ECIS 2024)*. AIS Electronic Library (AISeL). https://aisel.aisnet.org/ecis2024/track05_fow/track05_fow/3/

Hong, L., & Page, S. E. (2004). Groups of diverse problem solvers can outperform groups of high-ability problem solvers. *Proceedings of the National Academy of Sciences of the United States of America, 101*(46), 16385–16389. https://doi.org/10.1073/pnas.0403723101

Huang, L., Yu, W., Ma, W., Zhong, W., Feng, Z., Wang, H., Chen, Q., Peng, W., Feng, X., Qin, B., & Liu, T. (2025). A survey on hallucination in large language models: Principles, taxonomy, challenges, and open questions. *ACM Transactions on Information Systems, 43*(2), Article 42. https://doi.org/10.1145/3703155

Kalyan, A., Kumar, A., Chandrasekaran, A., Sabharwal, A., & Clark, P. (2021). How much coffee was consumed during EMNLP 2019? Fermi problems: A new reasoning challenge for AI. *Proceedings of the 2021 Conference on Empirical Methods in Natural Language Processing*, 7318–7328. Association for Computational Linguistics. https://doi.org/10.18653/v1/2021.emnlp-main.582

Khan, M. A., Ayub, U., Naqvi, S. A. A., Khakwani, K. Z. R., Sipra, Z. b. R., Raina, A., Zhou, S., He, H., Saeidi, A., Hasan, B., Rumble, R. B., Bitterman, D. S., Warner, J. L., Zou, J., Tevaarwerk, A. J., Leventakos, K., Kehl, K. L., Palmer, J. M., Murad, M. H., Baral, C., & Riaz, I. b. (2025).

Collaborative large language models for automated data extraction in living systematic reviews. *Journal of the American Medical Informatics Association, 32*(4), 638–647. https://doi.org/10.1093/jamia/ocae325

Kshetri, N., Dwivedi, Y. K., Davenport, T. H., & Panteli, N. (2024). Generative artificial intelligence in marketing: Applications, opportunities, challenges, and research agenda. *International Journal of Information Management, 75*, 102716. https://doi.org/10.1016/j.ijinfomgt.2023.102716

Lee, Y., Son, K., Kim, T. S., Kim, J., Chung, J. J. Y., Adar, E., & Kim, J. (2024). One vs. Many: Comprehending accurate information from multiple erroneous and inconsistent AI generations. In *Proceedings of the 2024 ACM Conference on Fairness, Accountability, and Transparency (FAccT '24)* (pp. 2518-2531). Association for Computing Machinery. https://doi.org/10.1145/3630106.3662681

Li, X., Wang, S., Zeng, S., Wu, Y., & Yang, Y. (2024). A survey on LLM-based multi-agent systems: Workflow, infrastructure, and challenges. *Vicinagearth, 1*, Article 9. https://doi.org/10.1007/s44336-024-00009-2

Lyon, A., Wintle, B. C., & Burgman, M. (2015). Collective wisdom: Methods of confidence interval aggregation. *Journal of Business Research, 68*(8), 1759–1767. https://doi.org/10.1016/j.jbusres.2014.08.012

Makanda, I. L. D., Jiang, P., Yang, M., & Shi, H. (2023). Emergence of collective intelligence in industrial cyber-physical-social systems for collaborative task allocation and defect detection. *Computers in Industry, 152*, 104006. https://doi.org/10.1016/j.compind.2023.104006

Memmert, L., & Tavanapour, N. (2023). Towards human-AI-collaboration in brainstorming: Empirical insights into the perception of working with a generative AI. In *Proceedings of the 31st European Conference on Information Systems (ECIS 2023)*. AIS Electronic Library (AISeL). https://aisel.aisnet.org/ecis2023_rp/429/

Memmert, L. (2024). Brainstorming with a generative language model: Understanding performance through brainstorming group effects. In *Proceedings of the 32nd European Conference on Information Systems (ECIS 2024)*. AIS Electronic Library (AISeL). https://aisel.aisnet.org/ecis2024/track06_humanaicollab/track06_humanaicollab/1/

Mienye, I. D., & Swart, T. G. (2025). Ensemble large language models: A survey. *Information, 16*(8), 688. https://doi.org/10.3390/info16080688

Mohammed, A., & Kora, R. (2023). A comprehensive review on ensemble deep learning: Opportunities and challenges. *Journal of King Saud University — Computer and Information Sciences, 35*(2), 757–774. https://doi.org/10.1016/j.jksuci.2023.01.014

Rosenberg, L. (2016). Artificial swarm intelligence vs. human experts. In 2016 International Joint Conference on Neural Networks (IJCNN) (pp. 2554–2561). IEEE. https://doi.org/10.1109/IJCNN.2016.7727517

Rosenberg, L., & Willcox, G. (2018). Artificial swarm intelligence vs. Vegas betting markets. In *2018 11th International Conference on Developments in eSystems Engineering (DeSE)* (pp. 55–60). IEEE. https://doi.org/10.1109/DeSE.2018.00014

Schumann, H., Willcox, G., Rosenberg, L., & Pescetelli, N. (2019). "Human swarming" amplifies accuracy and ROI when forecasting financial markets. In *2019 IEEE International Conference on Humanized Computing and Communication (HCC)* (pp. 73–78). IEEE. https://doi.org/10.1109/HCC46620.2019.00019

Simoiu, C., Sumanth, C., Mysore, A., & Goel, S. (2019). Studying the "Wisdom of Crowds" at Scale. *Proceedings of the AAAI Conference on Human Computation and Crowdsourcing, 7*(1), 171-179. https://doi.org/10.1609/hcomp.v7i1.5271

Storey, V. C., Yue, W. T., Zhao, J. L., & Lukyanenko, R. (2025). Generative artificial intelligence: Evolving technology, growing societal impact, and opportunities for information systems research. *Information Systems Frontiers.* https://doi.org/10.1007/s10796-025-10581-7

Taudien, A., Schiffels, S., Fuegener, A., Gupta, A., & Ketter, W. (2024). One versus many: Advice-taking from ensemble algorithms. In *Proceedings of the 32nd European Conference on Information Systems (ECIS 2024)*. AIS Electronic Library (AISeL). https://aisel.aisnet.org/ecis2024/track09_coghbis/track09_coghbis/9/

Vaswani, A., Shazeer, N., Parmar, N., Uszkoreit, J., Jones, L., Gomez, A. N., Kaiser, Ł., & Polosukhin, I. (2017). Attention is all you need. In *Advances in Neural Information Processing Systems (Vol. 30, pp. 5998–6008)*. https://doi.org/10.48550/arXiv.1706.03762

Wei, J., Wang, X., Schuurmans, D., Bosma, M., Ichter, B., Xia, F., Chi, E., Le, Q., & Zhou, D. (2022a). Chain-of-thought prompting elicits reasoning in large language models (Version 6). *arXiv*. https://doi.org/10.48550/arXiv.2201.11903

Wei, X., Zhang, Z., Zhang, M., Chen, W., & Zeng, D. D. (2022b). Combining crowd and machine intelligence to detect false news on social media. *MIS Quarterly, 46*(2), 977–1008. https://doi.org/10.25300/MISQ/2022/16526

Willcox, G., Rosenberg, L., Askay, D., Metcalf, L., Harris, E., & Domnauer, C. (2019). Artificial swarming shown to amplify accuracy of group decisions in subjective judgment tasks. Advances in Information and Communication: *Proceedings of the 2019 Future of Information and Communication Conference (FICC 2019)* (Lecture Notes in Networks and Systems, Vol. 70, pp. 373–383). Springer. https://doi.org/10.1007/978-3-030-12385-7_29

Willcox, G., Rosenberg, L., & Domnauer, C. (2020). The repeatability of human swarms. Advances in Information and Communication: *Proceedings of the 2020 Future of Information and Communication Conference (FICC 2020)* (Advances in Intelligent Systems and Computing, Vol. 1130, pp. 473–479). Springer. https://doi.org/10.1007/978-3-030-39442-4_35

Willcox, G., Rosenberg, L., Domnauer, C., & Schumann, H. (2021). Hyperswarms: A new architecture for amplifying collective intelligence. In *2021 IEEE 12th Annual Information Technology, Electronics and Mobile Communication Conference (IEMCON)* (pp. 207–212). IEEE. https://doi.org/10.1109/IEMCON53756.2021.9623239

Willcox, G., Rosenberg, L., & Schumann, H. (2023). Validating a new collective intelligence technology for accurate ranking using artificial swarm intelligence. *Proceedings of the Future Technologies Conference (FTC) 2023*, Volume 1 (Lecture Notes in Networks and Systems, Vol. 813, pp. 290–304). Springer. https://doi.org/10.1007/978-3-031-47454-5_22

Zhang, Y., Li, Y., Cui, L., Cai, D., Liu, L., Fu, T., Huang, X., Zhao, E., Zhang, Y., Chen, Y., Wang, L., Luu, A. T., Bi, W., Shi, F., & Shi, S. (2025). Siren's Song in the AI Ocean: A Survey on Hallucination in Large Language Models. *Computational Linguistics, 51*(1), 1–46. https://doi.org/10.1162/coli.a.16